%% file: example_paper.tex
\documentclass{article}

\usepackage{microtype}
\usepackage{graphicx}
\usepackage{caption}
\usepackage{subfigure}
\usepackage{booktabs} % for professional tables
\usepackage{algorithm}
\usepackage{algorithmic}
\usepackage{lineno}
\usepackage{epsfig}
\usepackage{comment}
\usepackage{multirow}
\usepackage{amsmath}
\usepackage{amsfonts}
\usepackage{xcolor}
\usepackage{adjustbox,booktabs}
\floatname{algorithm}{Algorithm}  
\renewcommand{\algorithmicrequire}{\textbf{Input:}}  
\renewcommand{\algorithmicensure}{\textbf{Output:}}

\usepackage{hyperref}

\usepackage[accepted]{icml2020}

\icmltitlerunning{Transfer Learning without Knowing: Reprogramming Black-box Machine Learning Models
}

\begin{document}

\twocolumn[
\icmltitle{Transfer Learning without Knowing: Reprogramming Black-box Machine Learning Models with Scarce Data and Limited Resources}




\begin{icmlauthorlist}
\icmlauthor{Yun-Yun Tsai}{nthu}
\icmlauthor{Pin-Yu Chen}{ibm}
\icmlauthor{Tsung-Yi Ho}{nthu}
\end{icmlauthorlist}

\icmlaffiliation{nthu}{National Tsing Hua University, Hsinchu, Taiwan}
\icmlaffiliation{ibm}{IBM Research}
\icmlcorrespondingauthor{Yun-Yun Tsai}{s107062548@m107.nthu.edu.tw}
\icmlcorrespondingauthor{Pin-Yu Chen}{pin-yu.chen@ibm.com}
\icmlcorrespondingauthor{Tsung-Yi Ho}{tyho@cs.nthu.edu.tw}
\icmlkeywords{Machine Learning, ICML}

\vskip 0.3in
]



\printAffiliationsAndNotice{}  % leave blank if no need to mention equal contribution

\begin{abstract}
Current transfer learning methods are mainly based on finetuning a pretrained model with target-domain data. Motivated by the techniques from adversarial machine learning (ML) that are capable of manipulating the model prediction via data perturbations, in this paper we propose a novel approach, black-box adversarial reprogramming (BAR), that repurposes a well-trained black-box ML model (e.g., a prediction API or a proprietary software) for solving different ML tasks, especially in the scenario with scarce data and constrained resources. The rationale lies in exploiting high-performance but unknown ML models to gain learning capability for transfer learning. Using zeroth order optimization and multi-label mapping techniques, BAR can reprogram a black-box ML model solely based on its input-output responses without knowing the model architecture or changing any parameter. More importantly, in the limited medical data setting, on autism spectrum disorder classification, diabetic retinopathy detection, and melanoma detection tasks, BAR outperforms state-of-the-art methods and yields comparable performance to the vanilla adversarial reprogramming method requiring complete knowledge of the target ML model. BAR also outperforms baseline transfer learning approaches by a significant margin, demonstrating cost-effective means and new insights for transfer learning.
\end{abstract}

\section{Introduction}
\label{sec:introduction}
\input{doc/introduction}

\section{Related Work}
\label{sec:relatedwork}
\input{doc/relatedwork}

\section{Black-box Adversarial Reprogramming (BAR): Method and Algorithm}
\label{sec:method}

\input{doc/method}

\section{Experiments}
\label{sec:experiment}
\input{doc/experiment}

\section{Conclusion}
\label{sec:conslusions}
\input{doc/conclusion}

\section*{Acknowledgements}
This work was based on a joint study agreement between IBM Research and National Tsing Hua University, Taiwan. The work of Yun-Yun Tsai and Tsung-Yi Ho was supported by the Taiwan Ministry of Science and Technology under Grant MOST 108-2218-E-007-031. Pin-Yu Chen would like to thank Payel Das at IBM Research for her inputs on the autism spectrum disorder classification task.

\nocite{langley00}

\bibliography{example_paper}
\bibliographystyle{icml2020}

\clearpage

\appendix
\label{sec:appendix}
\input{doc/supplementary}
%\section{Do \emph{not} have an appendix here}

%\textbf{\emph{Do not put content after the references.}}
%
%Put anything that you might normally include after the references in a separate supplementary file.

%%%%%%%%%%%%%%%%%%%%%%%%%%%%%%%%%%%%%%%%%%%%%%%%%%%%%%%%%%%%%%%%%%%%%%%%%%%%%%%
%%%%%%%%%%%%%%%%%%%%%%%%%%%%%%%%%%%%%%%%%%%%%%%%%%%%%%%%%%%%%%%%%%%%%%%%%%%%%%%

\end{document}

%% file: doc/introduction.tex
Transfer learning is a widely used practical machine learning (ML) methodology for learning to solve a new task in a target domain based on the knowledge transferred from a source-domain task \cite{pan2009survey}. One popular target-domain task is transfer learning of medical imaging with a large and rich benchmark dataset (e.g., ImageNet) as the source-domain task, since high-quality labeled medical images are often scarce and costly to acquire new samples \cite{raghu2019transfusion}. For deep learning models, transfer learning is often achieved by finetuning a pretrained source-domain model with the target-domain data, which requires complete knowledge and full control of the pretrained model, including knowing and modifying the model architecture and pretrained model parameters.

\begin{figure*}[t]
\centering
\includegraphics[width=0.99\linewidth]{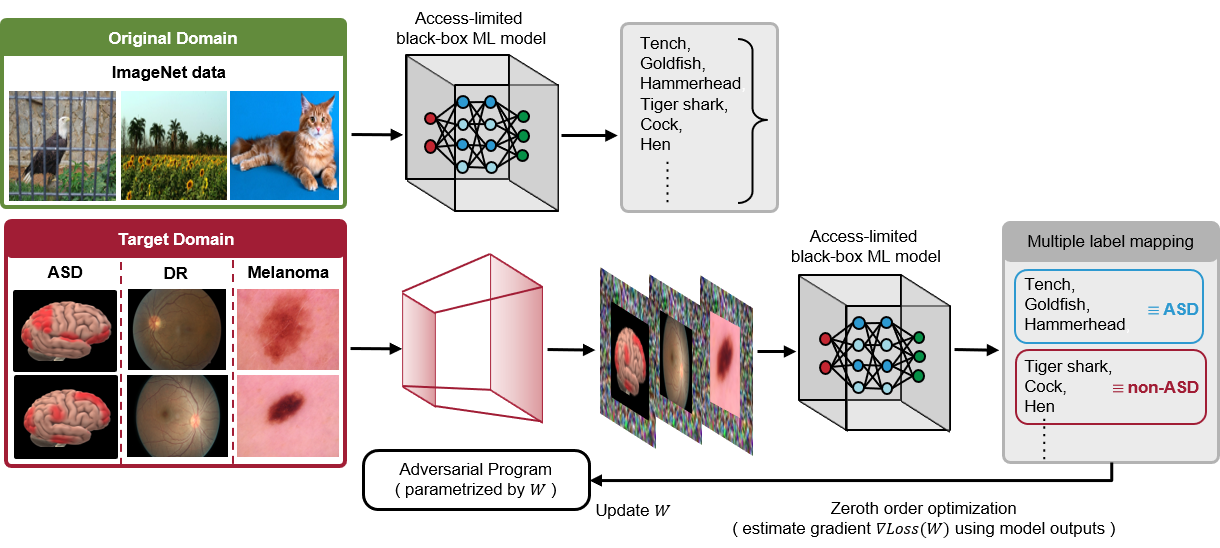}
%\vspace{-0.2in}
\vspace{-2mm}
\caption{Schematic overview of our proposed black-box adversarial reprogramming (BAR) method. }
\label{fig:overview}
%\vspace{-0.2in}
\end{figure*}
 
In this paper, we revisit transfer learning to address two fundamental questions: (i) Is finetuning a pretrained model necessary for learning a new task? (ii) Can transfer learning be expanded to \textit{black-box} ML models where nothing but only the input-output model responses (data samples and their predictions) are observable? In contrast, we call finetuning a \textit{white-box} transfer learning method as it assumes the source-domain model to be transparent and modifiable.

Recent advances in adversarial ML have shown great capability of manipulating the prediction of a well-trained deep learning model by designing and learning perturbations to the data inputs without changing the target model \cite{biggio2018wild}, such as prediction-evasive adversarial examples \cite{szegedy2013intriguing}. Despite of the ``vulnerability'' in deep learning models, these findings also suggest the plausibility of  transfer learning without modifying the pretrained model if  an appropriate perturbation to the target-domain data can be learned to align the target-domain labels with the pretrained source-domain model predictions. Indeed,
the \textit{adversarial reprogramming} (AR) method proposed in  \cite{elsayed2018adversarial} partially gives a negative answer to Question (i) by showing simply learning a universal target-domain data perturbation is sufficient to repurpose a pretrained source-domain model, where the domains and tasks can be different, such as reprogramming an ImageNet classifier to solve the task of counting squares in an image. However, the authors did not investigate the performance of AR on the limited data setting often encountered in transfer learning. Moreover, since the training of AR requires backpropagation of a deep learning model, AR still falls into the category of white-box transfer learning methods and hence does not address Question (ii).

To bridge this gap, we propose a novel approach, named black-box adversarial reprogramming (BAR), to reprogram a deployed ML model (e.g., an online image classification service) for black-box transfer learning. Comparing to the vanilla (white-box) AR approach, our BAR has the following substantial differences and unique challenges: 
\begin{enumerate}
    \item \textbf{Black-box setting.} The vanilla AR method assumes complete knowledge of the pretrained (target) model, which precludes the ability of reprogramming a well-trained but access-limited ML models such as prediction APIs or proprietary softwares that only reveal model outputs based on queried data inputs.
    \item \textbf{Data scarcity and resource constraint.} While data is crucial to most of ML tasks, in some scenarios such as medical applications, massive data collection can be expensive, if not impossible, especially when clinical trials, expert annotation or privacy-sensitive data are involved. Consequently,  without transfer learning, the practical limitation of data scarcity may hinder the strength of complex (large-scaled) ML models such as deep neural networks (DNNs). Moreover, even with moderate amount of data, researchers may not have sufficient computation resources or budgets to train a DNN as large as a commercial ML model or perform transfer learning on a large pretrained ML model. 
\end{enumerate}

Our proposed BAR tackles these two challenges in a cost-effective manner, which not only firstly extends white-box transfer learning to the black-box regime but also
``unlocks'' the power of well-trained but access-limited ML models for transfer learning. In particular, we focus on adversarial reprogramming of black-box image classification models for solving medical imaging tasks, as image classification is one of the most mature AI applications and many medical ML tasks often entail data scarcity challenges. As will be evident in the Experiments section (Sec. \ref{sec:experiment}), BAR can successfully leverage the powerful feature extraction capability of black-box ImageNet classifiers to achieve high performance in three medical image classification tasks  with limited data.

 Figure \ref{fig:overview} provides an overview of our proposed BAR method. To adapt to the black-box setting, we leverage zeroth-order optimization \cite{ghadimi2013stochastic} on iterative input-output model responses to enable black-box transfer learning. We also use multi-label mapping of source-domain and target-domain labels to enhance the performance of BAR. 
 %To address data scarcity and resource constraint, we show that BAR can  maintain similar performance while reducing training data size. Note that the effect of data scarcity and resource constraint has not been studied in the original AR work.
%We also report the low expense of implementing BAR using online computation resources such as Amazon Web Service (AWS). 
%For instance, BAR can reprogram an online image classifier for classifying autism spectrum disorder with \textcolor{red}{62.30\%} accuracy while merely costing \textcolor{red}{\$23.94} US dollars on AWS. 
We summarize our main contributions as follows.

\begin{itemize}
    \item We propose BAR, a novel approach to reprogram black-box ML models for transfer learning. To the best of our knowledge, BAR is the first work that expands transfer learning to the black-box setting without knowing or finetuning the pretrained model. 
    \item We evaluate the performance of BAR using three different medical imaging tasks for transfer learning from pretrained ImageNet models: (a)  autism spectrum disorder (ASD) classification; (b) diabetic retinopathy  (DR) detection; and (c) melanoma detection. The results show that our method consistently outperforms the state-of-the-art methods and improves the accuracy of the finetuning approach by a significant margin. 
    We also explain the success of BAR through a representation analysis and several ablation studies.
    \item  We demonstrate the practicality and effectiveness of BAR by reprogramming real-life image classification APIs from Clarifai.com\footnote{\url{https://www.clarifai.com}} and Microsoft Custom Vision\footnote{\url{https://azure.microsoft.com/en-us/services/cognitive-services/custom-vision-service/}}, which is infeasible for the vanilla white-box AR method due to the black-box setting. In terms of total expenses, it only costs   
    less than \$24 US dollars to reprogram these two APIs for ASD classification.
    %\textcolor{red}{For the price of building and running a Machine Learning Model, it only costs \$0.900 per hour hosted on Amazon Web Service (AWS) p2.xlarge \footnote{\url{https://aws.amazon.com/ec2/instance-types/p2/?nc1=h_ls}} instance. For the price of querying, it takes \$23.04 US dollars. Our result shows that we only need \$23.94 US dollars with 1500 data samples to reprogram the Clarifai API for autism spectrum disorder (ASD) classification task and can achieves 62.30\% classification accuracy which is close to state-of-the-art results.}
\end{itemize}

%% file: doc/relatedwork.tex
\subsection{Adversarial ML and Reprogramming}
Adversarial ML mainly studies how to manipulate the decision-making of a target model and develop countermeasures \cite{biggio2018wild}. In particular, several works have identified the vulnerability of DNNs to different types of adversarial threats, such as crafting prediction-evasive adversarial examples \cite{biggio2013evasion,szegedy2013intriguing,goodfellow2014explaining,carlini2017towards,chen2017ead} or poisoning training data and implanting backdoors in the downstream ML models \cite{munoz2017towards,chen2017targeted,shafahi2018poison,BadNet_Access,xie2020dba}, to name a few. 

Adversarial reprogramming (AR) is a recently introduced technique that aims to reprogram a target ML model for performing a different task \cite{elsayed2018adversarial}. %For instance, a number of DNN models trained for ImageNet classification task were successfully reprogrammed for solving the task of counting squares in an image.  
Different from typical transfer learning methods that modify the model architecture or parameters for solving a new task with target-domain data, AR keeps the model architecture unchanged. Instead, AR uses a trainable adversarial program and a designated output label mapping on the target-domain data samples to perform reprogramming. Intuitively, the adversarial program serves as a parametrized and trainable input transformation function such that when applied to the target-domain data  (e.g., images having squares), the \textit{same} target model will be reprogrammed for the new task (e.g., the output label ``dog'' of a programmed data input translates to ``3 squares''). The work in \cite{neekhara2018adversarial} demonstrates AR of text classification but still assumes white-box access to the target ML models.

\subsection{Zeroth Order Optimization for Black-box Setting}
It is worth noting that the vanilla AR method proposed in \cite{elsayed2018adversarial} requires complete access to the target ML model to allow back-propagation for training the parameters of adversarial program. In other words, vanilla AR lacks the ability to reprogram an access-limited ML model such as prediction API, owing to prohibited access to the target model disallowing back-propagation. To bridge this gap and empower reprogramming advanced yet access-limited ML models trained with tremendous amount of data and considerable computation resources (e.g., Google cloud vision API), we use zeroth order optimization techniques to enable black-box AR for transfer learning.

In contrast to the conventional first order (gradient-based) optimization methods such as stochastic gradient descent (SGD),
zeroth order optimization \cite{ghadimi2013stochastic} achieves gradient-free optimization by merely using numerical evaluations of the same training loss function instead of gradients, making it a powerful tool for the black-box setting. As the gradients of black-box ML models are infeasible to obtain, the main idea of zeroth order optimization is to replace 
the true gradients in first-order algorithms with
gradient estimates from function evaluations.
Prior arts in adversarial ML have shown that zeroth order optimization can be used to generate adversarial examples \cite{chen2017zoo,tu2018autozoom,brendel2017decision,ilyas2018black,cheng2018query,cheng2020signopt}, known as black-box adversarial attacks. Moreover, advanced zeroth order optimization methods can provide query-efficient solutions for black-box ML tasks \cite{liu2018zeroth,liu2018signsgd,liu2020primer}.

%% file: doc/method.tex
This section presents our proposed method and algorithm, named BAR, for reprogramming black-box ML models. A schematic overview of BAR is illustrated in Figure \ref{fig:overview}.

\subsection{Problem Formulation}
\textbf{Black-box setting}. We consider the problem of reprogramming a black-box ML classification model denoted by $F:\mathcal{X} \mapsto \mathbb{R}^K$, where it takes a data sample $X \in \mathcal{X}$ as an input and gives a vector of confidence scores $F(X)=[F_1(X),F_2(X),\ldots,F_K(X)] \in \mathbb{R}^K$ as its output, where $\mathcal{X}$ denotes the space of feasible data samples (e.g., image sizes and pixel value ranges) and $K$ is the number of classes. Similar to the access rights of a regular user when using a prediction API, one is able to observe the model output $F(X)$ for any given $X \in \mathcal{X}$, whereas inquiring the gradient $\nabla F(X)$ is inadmissible.

%In our work, we consider an adversary is unable to access to the parameters of target black-box model but still have to re-purpose the target classification task. The only information that adversary can require is output results from the last softmax layer of model. 
\textbf{Adversarial program}. To reprogram a black-box ML model, we use the same form of adversarial program in \cite{elsayed2018adversarial} as an input transformation function to translate the data of the target domain to the input space of the source domain.
Without loss of generality, let $\mathcal{X} = [-1,1]^d$ denote the scaled input space of an ML model $F$, where $d$ is the (vectorized) input dimension. We also denote the set of data from the target domain by $\{D_i\}_{i=1}^n$, where $D_i \in [-1,1]^{d^\prime}$ and $d^\prime < d$ to allow extra dimensions for reprogramming. For each data sample $i \in [n]$, where $[n]$ denotes the integer set  $\{1,2,\ldots,n\}$,  we let $X_i$ be the zero-padded data sample containing $D_i$, such as embedding a brain-regional correlation graph of size 200 $\times$ 200 to the center of a 299 $\times$ 299 (width $\times$ height) image, as shown in Figure \ref{fig:overview}. Let $M \in \{0,1\}^d$ be a binary mask function indicating the common embedding location for $\{D_i\}_{i=1}^n$, where $M_j = 0$ means the $j$-th dimension is used for embedding and $M_j = 1$ otherwise. 
The transformed data sample for AR is defined as
\begin{align}
\label{eqn_adv_program}
   \widetilde{X}_i = X_i + P~~\text{and}~~  P = \tanh(W \odot M),
\end{align}
where $P$ is called an adversarial program to be learned and is universal to all target data samples $\{X_i\}_{i=1}^n$, $W \in \mathbb{R}^d$ is a set of trainable parameters for AR, $\odot$ denotes the Hadamard (entry-wise) product, and $\tanh \in [-1,1]$ ensures $ \widetilde{X}_i \in [-1,1]^d$. Note that the binary mask function $M$ in $P$ ensures the target data samples $\{D_i\}$ embedded in $\{\widetilde{X}_i\}$ are intact during AR. 

\textbf{Multi-label mapping (MLM).} As illustrated in Figure \ref{fig:overview}, in addition to input data transformation via an adversarial program, for AR we also need to map the source task's output labels (e.g., different objects) to the target task's output labels (e.g., ASD or non-ASD). Evaluated on three medical tasks (see Section \ref{sec:experiment}), we find that multiple-source-labels to  one-target-label mapping can further improve the accuracy of the target task when compared to one-to-one label mapping. For instance, the prediction of a transformed data input from the source label set \{Tench,Goldfish,Hammerhead\} will be reprogrammed for predicting the target class ASD. 
Let $K$ $(K^\prime)$ be the total number of classes for the source (target) task. We use the notation $h_j(\cdot)$ to denote the $k$-to-1 mapping function that averages the predictions of a group of $k$ source labels as the prediction of the $j$-th target domain's label. For example, If the source labels \{Tench,Goldfish,Hammerhead\} map to the target label \{ASD\}, then $h_{\text{ASD}}(F(X))=[F_{\text{Tench}}(X)+F_{\text{Goldfish}}(X)+F_{\text{Hammerhead}}(X)]/3$. More generally, if a subset of source labels $\mathcal{S} \subset [K]$ map to a target label $j \in [K^\prime]$, then $h_j(F(X))=\frac{1}{|S|}\sum_{s \in S} F_s(X)$, where $|\mathcal{S}|$ is the set cardinality. 
Furthermore, we propose a frequency-based label mapping scheme by matching target labels to source labels according to the label distribution of initial predictions on the target-domain data before reprogramming. We find that it improves the accuracy over random label mapping.
We defer the readers to Section \ref{sec:experiment} for more details and an ablation study on multi-label mapping in Sec. \ref{subsec_ablation}.
%\footnote{Also see a blog post of multi-label mapping on other tasks \text{https://rajatvd.github.io/Exploring-Adversarial-Reprogramming}.}

%Define the adv labels for training loss.

%is the adversarial program parameters to be learned. M is the mask that we want to add on the input image to mask out the central part of adversarial image. Note that we will put our original image at the center and pad the perturbation outside. We use tanh (·) to bound the adversarial perturbation to be in (−1, 1) – the same range as the rescaled ImageNet images the target networks are trained to classify. The adversarial images can be defined as below

%Unlike other adversarial perturbation, we don't need to constraint the magnitude  of perturbation P in adversarial program. Besides, this adversarial program is universal to all the images, not specific to individual image.
%After transforming our input data, the size of adversarial input data will consistent with the input shape of black box model. 
%We then use the adversarial input to query our black box model and do the label mapping. As our model output have 1000 classes, we need to map it as 10 classes to classify Mnist data. 

\textbf{Loss function for AR.}
%\textcolor{blue}{still need to be revised. a little weird}
Here we formally define the training loss for AR. Without loss of generality, we assume the model output is properly normalized such that $\sum_{j=1}^K F_j(X)=1$ and $ F_j(X) \geq 0$ for all $j \in [K]$, which can be easily satisfied by applying a softmax function to the model output.
Let $\{y_i\}_{i=1}^n$ with $y_i=[y_{i1},\ldots,y_{iK^\prime}] \in \{0,1\}^{K^\prime}$ denote the one-hot encoded label for the target domain task and let $h(F(X))=[h_1(F(X)),\ldots,h_{K^\prime}(F(X))]$ be a surjective multi-label mapping function from the model prediction $F(X)$ of the source domain to the target domain. For training the adversarial program $P$ parametrized by $W$, we use the focal loss \cite{lin2017focal} as empirically it can further improve the performance of AR/BAR (see Sec. \ref{subsec_ablation}). The focal loss (F-loss) aims to penalize the samples having low prediction probability during training, and it includes the conventional cross entropy loss (CE-loss) as a special case. The focal loss of the ground-truth label $\{y_i\}_{i=1}^n$ and the transformed prediction probability $\{h \left( F(X_i+P) \right)\}_{i=1}^n$ is
\begin{align}
\label{eqn:loss_func}
      -\sum_{i=1}^{n} \sum_{j=1}^{K^\prime} \omega_j (1-h_j)^\gamma y_{ij} \log h_j \left( F(X_i+P) \right),
\end{align}
where $\omega_j > 0$ is a class balancing coefficient, $\gamma \geq 0$ is a focusing parameter which down-weights high-confidence (large $h_j$) samples. When $\omega_j=1$ for all $j$ and $\gamma=0$, the focal loss reduces to the cross entropy. In our implementation, we set $\omega_j= 1/n_j$ and $\gamma=2$, where $n_j$ is the number of samples in class $j$, as suggested in \cite{lin2017focal}.

%$\alpha$ is a weighting factor which can balance the importance of positive/negative example in binary case. It slightly improves accuracy over the non-$\alpha$-balanced form.
% \begin{align}
% \label{eqn:loss_func}
%         CE-Loss=-\sum_{i=1}^{n} \sum_{j=1}^{K^\prime} y_{ij} \log h_j \left( F(X_i+P) \right).
% \end{align}

% \begin{align}{}
% \label{eqn:focal_loss_func}
%     &F-Loss= -\alpha\cdot(1-p)^\gamma\cdot \log p, \\
%     &p = \sum_{i=1}^{n} \sum_{j=1}^{K^\prime} y_{ij} h_j \left( F(X_i+P) \right)
% \end{align}

% \begin{align*}
%     p = \sum_{i=1}^{n} \sum_{j=1}^{K^\prime} y_{ij} h_j \left( F(X_i+P) \right)
%     %\left\{
%     %            \begin{array}{@{}ll@{}}
%     %              p, & \text{if}\ y_{ij}=1 \\
%     %              1-p, & 0,\text{otherwise}
%     %            \end{array}\right.
% \end{align*}

Note that the loss function is a function of $W$ since 
 from \eqref{eqn_adv_program} the adversarial program $P$ is parametrized by $W$,  and $W$ is the set of optimization variables to be learned for AR. The loss function can be further generalized to the minibatch setting for stochastic optimization.

\subsection{Zeroth Order Optimization for BAR}
%\subsection{Random Vector based Gradient Estimation}
In the white-box setting assuming complete access to the target ML model $F$, optimizing the loss function in \eqref{eqn:loss_func} and retrieving its gradient for AR are straightforward via back-propagation. However, when $F$ is a black-box model and only the model outputs $F(\cdot)$ are available for AR, back-propagation through $F$ is infeasible since the gradient $\nabla F(\cdot)$ is inadmissible. In our BAR framework, to optimize the loss function in \eqref{eqn:loss_func} and update the parameters $W$ of the adversarial program, we propose to use zeroth order optimization to solve for $W$. Specifically, there are two major components to enable BAR: (i) gradient estimation and (ii) gradient descent with estimated gradient.

\textbf{Query-efficient gradient estimation.} Let $f(W)$ be the $Loss$ defined in \eqref{eqn:loss_func} and $W$ be the optimization variables. To estimate the gradient $\nabla f(W)$, we use the one-sided averaged gradient estimator \cite{liu2018zeroth,tu2018autozoom} via $q$ random vector perturbations, which is defined as 
\begin{align}
\label{eqn_avg_gradient_est}
        \bar{g}(W) = \frac{1}{q} \sum_{j=1}^{q} g_{j},
\end{align}
where $\{g_j\}_{j=1}^q$ are $q$ independent random gradient estimates of the form
\begin{align}
\label{eqn_single_graidnt_est}
    g_j = b \cdot \frac{f(W+\beta U_j)-f(W)}{\beta} \cdot U_j,
\end{align}
where $b$ is a scalar balancing bias and variance trade-off of the estimator, $W \in \mathbb{R}^d$ is the set of optimization variables in vector form, $\beta$ is the smoothing parameter, and $U_j \in \mathbb{R}^d$ is a vector that is uniformly drawn at random from a unit Euclidean sphere. The mean squared estimation error between $\bar{g}(W)$ and the true gradient $\nabla f(W)$ has been characterized in \cite{tu2018autozoom} with mild assumptions on $f$.  In our experimental setup, we set $b=d$ in order to obtain an unbiased gradient estimator of a smoothed function of $\nabla f$ \cite{gao2014information}, and
we set $\beta$ to be of the order $1/d$ (i.e., $\beta=0.01$) following the analysis in \cite{liu2018zeroth} and set $U_j$ to be a realization of a standard normal Gaussian random vector divided by its Euclidean norm. By construction, for each data sample $X_i,~i \in [n]$, the averaged gradient estimator takes $q+1$ queries from the ML model $F$. Smaller $q$ can reduce the number of queries to the target model but may incur larger gradient estimator error. We will study the influence of $q$ on the performance of BAR in the next section.

\textbf{BAR algorithm.}
Using the averaged gradient estimator $\bar{g}$, our BAR algorithm is compatible with any gradient-based training algorithm by simply replacing the inadmissible gradient $\nabla Loss$ with $\bar{g}$ in the gradient descent step. The corresponding algorithmic convergence guarantees have been proved in recent works such as \cite{liu2018zeroth,liu2018signsgd} in both the convex loss and non-convex loss settings. In this paper, we use stochastic gradient descent (SGD) with $\bar{g}$ to optimize the parameters $W$ in BAR, which are updated by
\begin{align}
\label{eqn:gradient_update}
    W_{t+1} = W_t - \alpha_t \cdot   \bar{g}(W_t),
\end{align}
where $t$ is the $t$-th iteration for updating $W$ with a minibatch sampled from $\{X_i\}_{i=1}^n$ (we set the minibatch size to be 20),
$\alpha_t$ is the step size (we use exponential decay with initial learning rate $\eta$), and $\bar{g}(W_t)$ is the gradient estimate of the loss function at $W_t$ using the $t$-th minibatch.
Note that since the loss function defined in \eqref{eqn:loss_func} is a function of the target ML model $F$'s input and output, and the parameters $W$ of the adversarial program only associate with the input of $F$, the entire gradient estimation and training process for BAR is indeed operated in a black-box manner. That is, BAR only uses input-output responses of $F$ and does \textit{not} assume access to the model internal details such as model type, parameters, or source-domain data.
The entire training process for BAR takes $\#~iterations \times mini~batch~ size \times~(q+1)$ queries to $F$. Algorithm \ref{algo_dnn_training} summarizes our proposed BAR method. For the ease of description, the minibatch size is set to be the training data size $n$ in Algorithm \ref{algo_dnn_training}.
% We propose 

%\subsection{Algorithm}
\begin{algorithm}
    \begin{algorithmic}[1]
    \caption{Training algorithm of black-box adversarial reprogramming (BAR)}
        \label{algo_dnn_training}
            \REQUIRE {black-box ML model $F$, AR loss function $Loss(\cdot)$, target domain training data $\{D_i,y_i\}_{i=1}^n$, maximum number of iterations $T$, number of random vectors for gradient estimation $q$, multi-label mapping function $h(\cdot)$, step size $\{\alpha_t\}_{t=1}^T$ }
            \ENSURE  {Optimal adversarial program parameters $W$ }
            \STATE Randomly initialize $W$; set $t=1$ \STATE Embed $\{D_i\}_{i=1}^n$ with mask $M$ to create $\{X_i\}_{i=1}^n$
            \WHILE{$t \leq T$}
                \STATE \# \textbf{Generate adversarial program}\\
                 %Apply random gradient estimator in (4) and get a set $W'= \{W_{0}, W_{1}, W_{2}, ..., W_{q+1}\}$, where q=5.\\
                 $P= \tanh(W \odot M)$ \\
                 % P is a perturbation set
                \# \textbf{Generate $q$ perturbed adversarial programs}\\
                $\widetilde{P}_j= \tanh((W + U_j) \odot M)$ for all $j \in [q]$ \\$\{U_j\}_{j=1}^q$ are random vectors defined in \eqref{eqn_single_graidnt_est}
                 \STATE
                 \# \textbf{Function evaluation for gradient estimation}
                 
                 Evaluate $Loss$ in \eqref{eqn:loss_func} with $W$ and $\{X_i+P\}_{i=1}^n$\\
                 Evaluate $Loss$ in \eqref{eqn:loss_func} with $W+U_j$ and $\{X_i+\widetilde{P}_j\}_{i=1}^n$ for all $j \in [q]$ \\                 
                 %$S_{j} = \{X_{j}+P_{0}, X_{j}+P_{1}, ..., X_{j}+P_{q+1}\}$\\
                 \STATE \# \textbf{Optimize adversarial program's parameters:}\\
                 %$L' \gets \{( Loss(F(x')), x') | x' \in S \}$\\
                 Use Step 5 and  \eqref{eqn_avg_gradient_est} to obtain estimated gradient $\bar{g}(W)$\\
                 $W \gets W - \alpha_t \cdot \bar{g}(W)$\\
                 $t \gets t+1$
                 
            \ENDWHILE
    \end{algorithmic}
\end{algorithm}

% Since we are target on black box model, it is unable to use back propagation to update our weight with loss. The authors in (Chen et al. 2017) use the symmetric difference quotient method (Lax and Terrell 2014) to evaluate the gradient $\frac{\partial f(x)}{\partial x_{i}}$ of the i-th component. Inspired by them, We estimate our gradient by using random vector based gradient estimation. The formulation is shown below.\\
% \begin{equation}
% \small
%     g = b \cdot \frac{f(W+\beta u)-f(W)}{\beta} \cdot u
% \label{algorithm:grad}
% \end{equation} 
% where $\beta>0$ is a smoothing parameter. $W$ is the weight parameter in transform function that we need to update. $u$ is a unit-length vector that is uniformly drawn at random from a unit Euclidean sphere, and $b$ is a tunable scaling parameter that balances the bias and variance trade-off of the gradient estimation error.
% We first generate several weight with unit random vectors and apply them on original dataset. $g$ is the estimate gradient, which is the estimate result of $\partial loss$ / $\partial W$. To effectively control the error in gradient estimation, we consider a more
% general gradient estimator, in which the gradient estimate is averaged over $q$ random directions $\left\{ u_{j} \}_{j=1}^q$. That is,
% \begin{equation}
% \small
%     \bar{g} = \frac{1}{q} \sum_{j=1}^{q} g_{j}
% \label{algorithm:average_grad}
% \end{equation} 
% where $g_{j}$ is a gradient estimate defined in (4) with $u = u_{j}$.
% The use of multiple random directions can reduce the variance
% of $g$ in (4) for convex loss functions.

%% file: doc/experiment.tex
This section presents the following experiments for performance evaluation and comparison. 
\begin{enumerate}
    \item Reprogramming three pretrained black-box ImageNet classifiers (1000-object recognition task) from \cite{model_link}, including ResNet 50 \cite{He2015DeepRL}, Inception V3 \cite{Szegedy2015RethinkingTI} and DenseNet 121 \cite{iandola2014densenet},
    for three medical imaging classification tasks, including Autism Spectrum Disorder (ASD) classification (2-classes), Diabetic Retinopathy (DR) detection (5-classes) and Melanoma detection (7-classes).
    \item Reprogramming two online Machine Learning-as-a-Service (MLaaS) toolsets, including Clarifai.com\footnotemark[1] and Microsoft Custom Vision\footnotemark[2], for medical imaging tasks and reporting the expenses.
    \item Sensitivity analysis on the influence of number of random vectors $q$ and multi-label mapping (MLM) size $m$ for BAR, and ablation studies in terms of different loss functions (CE-loss v.s. F-loss) and label mapping methods (random mapping v.s. frequency mapping). 
\end{enumerate}

For implementing BAR and AR, we use the focal loss in \eqref{eqn:loss_func} and frequency-based MLM derived from the initial predictions of the target-domain data before reprogramming.
Their ablation studies will be discussed in Section \ref{subsec_ablation}. We also highlight the results of BAR in boldface.

%\textcolor{blue}{Need to mention MLM, minibatch size, $q$ and other parameters in each subsection}

\textbf{Baselines.} To benchmark the performance of BAR, we compare it with three baselines. For fair comparisons, all methods use the same training/testing data and 
we do not use any data augmentation nor model ensemble techniques.
\begin{itemize}
     \item Vanilla adversarial reprogramming (white-box AR): It assumes white-box access to the target ML model and optimizes the AR training loss in \eqref{eqn:loss_func} using the ADAM optimizer \cite{kingma2014adam}. Its accuracy serves as an upper bound of BAR as BAR only assumes black-box access. %But we observe that in some cases BAR's accuracy can be slightly better than white-box AR.
    \item Transfer learning: We finetune the same pretrained models following the implementation in tensorflow tutorial\footnote{\url{https://www.tensorflow.org/tutorials/images/transfer_learning}}. The details are given in the supplementary material.
    We also implement another baseline that trains the model from scratch. Ideally, in the limited data setting training from scratch serves as a lower bound on BAR's accuracy due to insufficient data. We use the original target-domain data (without zero padding) for these two transfer learning baselines because the resulting performance is better than  that with zero padding.
    \item State-of-the-art (SOTA): For each task, we implement the SOTA methods in the literature but disable any data augmentation or model ensemble techniques. 
\end{itemize}

%\begin{table}[t]
%\centering
%\footnotesize
%\caption{Performance comparison on face recognition.}
%\vspace{1.5mm}
%\setlength{\tabcolsep}{3pt}
%The testing data size is 150}
%\adjustbox{max width=1\linewidth}{
%\begin{tabular}{ccccc}
%\hline
%\textbf{Model}  & \textbf{\begin{tabular}[c]{@{}c@{}}training\\ %size\end{tabular}} & \textbf{\begin{tabular}[c]{@{}c@{}}cnn\\ %Acc.\end{tabular}} & \textbf{\begin{tabular}[c]{@{}c@{}}white box\\ %Acc.\end{tabular}} & \textbf{\begin{tabular}[c]{@{}c@{}}black box\\ %Acc.\end{tabular}} \\ \hline
%\multirow{3}{*}{\textbf{\begin{tabular}[c]{@{}c@{}}Restnet 50\\with %MLM\end{tabular}}}    
%& 200 & 90.55\% & 93.21 \% & 92.86\%  \\ 
%& 300 & 90.21\% & 94.52 \% & 94.29\%  \\
%& 400 & 91.50\% & 96.10\% & 95.71\%   \\  \hline
%\multirow{3}{*}{\textbf{\begin{tabular}[c]{@{}c@{}}Incept.V3\\ with %MLM\end{tabular}}} 
%& 200 & 90.55\% & 91.18\% & 90.71\%   \\
%& 300 & 91.21\% & 91.25\% & 91.33\%    \\
%& 400 & 91.50\% & 93.25\% & 92.36\%   \\ \hline  
%\end{tabular}}
%\label{table:face}
%\end{table}

\subsection{Autism Spectrum Disorder Classification}
Classifying Autism Spectrum Disorder (ASD) is a challenging task. ASD is a complex developmental disorder that involves persistent challenges in social interaction, speech and nonverbal communication, and restricted/repetitive behaviors. It affects about 1\% of the global population. Currently, the only clinical method for diagnosing ASD are standardized ASD tests, which require prolonged diagnostic time and considerable medical costs. Therefore, ML can play an important role in providing cost-effective means of detecting ASD.
We use the dataset from the Autism Brain Imaging Data Exchange (ABIDE) database \cite{abide}. 
%ABIDE is an international collaborative project that collected data from a large number of ASD patients and typical non-ASD controls from 17 research and clinical institutes. 
The preprocessed dataset\footnote{\url{http://preprocessed-connectomes-project.org/abide}} is split into 10 folds and contains 503 individuals suffering from ASD and 531 non-ASD samples. The data sample is a 200 $\times$ 200 brain-regional correlation graph of fMRI measurements, which is embedded in each color channel of ImageNet-sized inputs. In this task, we assign 5 separate ImageNet labels to each ASD label (i.e., ASD/non-ASD) for MLM and
set the parameters $\eta=0.05$ and $q=25$.
Table \ref{table:abide} reports the 10-fold cross validation test accuracy, where the averaged test data size is 104.
The accuracy of BAR is comparable to white-box AR, and their accuracy outperforms the SOTA performance as reported in \cite{Heinsfeld2018Identification, Eslami_2019}. The performance of finetuing and training from scratch is merely close to random guessing due to limited data, and BAR's accuracy is 17\%-18\% better than that of transfer learning.

%\begin{table}[]
%\centering
%\footnotesize
%\setlength{\tabcolsep}{3pt}
%\caption{Performance comparison (10-fold averaged test accuracy) on autism spectrum disorder classification task. }
%\vspace{1.5mm}
%\adjustbox{max width=1\linewidth}{
%\begin{tabular}{ccccc}
%\hline
%\textbf{Model}  & \textbf{\begin{tabular}[c]{@{}c@{}}training\\ size %(avg.)\end{tabular}} & \textbf{\begin{tabular}[c]{@{}c@{}}cnn\\ %Acc.\end{tabular}} & \textbf{\begin{tabular}[c]{@{}c@{}}white box\\ %Acc.\end{tabular}} & \textbf{\begin{tabular}[c]{@{}c@{}}black box\\ %Acc.\end{tabular}} \\ \hline
%\multirow{3}{*}{\textbf{\begin{tabular}[c]{@{}c@{}}Restnet 50\\with %MLM\end{tabular}}}    
%& 230  & 50.00\%    & 56.80\%  & 54.18\%   \\  
%& 465  & 48.71\%    & 62.55\%  & 57.00\%   \\  
%& 930  & 52.99\%    & 62.03\%  & 62.13\%   \\ \hline
%\multirow{3}{*}{\textbf{\begin{tabular}[c]{@{}c@{}}Incept.V3\\ with %MLM\end{tabular}}} 
%& 230  & 50.00\%   &  60.12\%  & 57.55\% \\  
%& 465 & 48.71\%   &   62.14\%  & 60.21\%  \\   
%& 930 & 52.99\%    &  65.00\%  & 61.15\% \\ \hline
%\end{tabular}}
%\label{table:abide}
%%\end{table}

\begin{table}[]
\centering
%\footnotesize
%\setlength{\tabcolsep}{3pt}
\caption{Performance comparison (10-fold averaged test accuracy) on autism spectrum disorder classification task.}
\vspace{1.5mm}
\adjustbox{max width=1\linewidth}{
\begin{tabular}{cccccc}
\hline
\textbf{Model}  & \textbf{\begin{tabular}[c]{@{}c@{}}Accuracy\end{tabular}} & \textbf{\begin{tabular}[c]{@{}c@{}}Sensitivity\end{tabular}} & \textbf{\begin{tabular}[c]{@{}c@{}}Specificity\end{tabular}} \\ \hline
\multirow{1}{*}{\textbf{\begin{tabular}[c]{@{}c@{}}ResNet 50 (BAR) \end{tabular}}}    
%& 465  & 67.63\%    & 65.55\%  & 62.00\%   \\  
 & \textbf{70.33}\%    & \textbf{69.94}\%  & \textbf{72.71}\% \\ 
\multirow{1}{*}{\textbf{\begin{tabular}[c]{@{}c@{}}ResNet 50 (AR) \end{tabular}}}    
%& 465  & 67.63\%    & 65.55\%  & 62.00\%   \\  
& 72.99\%    & 73.03\%  & 72.13\% \\ 
 \multirow{1}{*}{\textbf{\begin{tabular}[c]{@{}c@{}} Train from scratch \end{tabular}}} 
 %& 50.96\%   &  50.13\%  & 52.34\% \\ 
 & 51.55\%   &  51.17\%  & 53.56\% \\ 
\multirow{1}{*}{\textbf{\begin{tabular}[c]{@{}c@{}}Transfer Learning (finetuned)\end{tabular}}} 
 & 52.88\%   &  54.13\%  & 54.70\% \\ \hline
\multirow{1}{*}{\textbf{\begin{tabular}[c]{@{}c@{}}Incept. V3 (BAR) \end{tabular}}}    
%& 465  & 67.63\%    & 65.55\%  & 62.00\%   \\  
 & \textbf{70.10}\%   &  \textbf{69.40}\%  & \textbf{70.00}\%  \\ 
\multirow{1}{*}{\textbf{\begin{tabular}[c]{@{}c@{}}Incept. V3 (AR)\end{tabular}}} 
%& 465 & 65.71\%   &  66.14\%  & 65.21\% \\   
 & 72.30\%    & 71.94\%  & 74.71\%  \\ 
\multirow{1}{*}{\textbf{\begin{tabular}[c]{@{}c@{}} Train from scratch \end{tabular}}} 
% & 49.80\%   &  50.40\%  & 51.55\% \\ 
 & 50.20\%   &  51.43\%  & 52.67\% \\ 
\multirow{1}{*}{\textbf{\begin{tabular}[c]{@{}c@{}}Transfer Learning (finetuned)\end{tabular}}} 
 & 52.10\%   &  52.65\%  & 54.42\% \\ \hline
\multirow{1}{*}{\textbf{\begin{tabular}[c]{@{}c@{}}SOTA 1. \cite{Heinsfeld2018Identification} \end{tabular}}} 
 & 65.40\%   &  69.30\%  & 61.10\% \\ \hline
\multirow{1}{*}{\textbf{\begin{tabular}[c]{@{}c@{}}SOTA 2. \cite{Eslami_2019} \end{tabular}}} 
 & 69.40\%   &  66.40\%  & 71.30\% \\ \hline
\end{tabular}}
\label{table:abide}
\vspace{-4mm}
\end{table}

\begin{table}[]
\centering
%\footnotesize
%\setlength{\tabcolsep}{4pt}
\caption{Test accuracy on diabetic retinopathy detection task.  The notation $*$ denotes the network used in SOTA method.}
\vspace{1.5mm}
\adjustbox{max width=1\linewidth}{
\begin{tabular}{cccccc}
\hline
\textbf{Model}  & \textbf{\begin{tabular}[c]{@{}c@{}}From Scratch\end{tabular}}& \textbf{\begin{tabular}[c]{@{}c@{}}Finetuning\end{tabular}} & \textbf{\begin{tabular}[c]{@{}c@{}}AR \end{tabular}} & \textbf{\begin{tabular}[c]{@{}c@{}}BAR \end{tabular}} \\ \hline
\multirow{1}{*}{\textbf{\begin{tabular}[c]{@{}c@{}}ResNet 50*  \end{tabular}}}    
%& 800  &  & 71.84\%    &  72.00\%  & 71.46\%   \\  
%& 1500 &  & 72.62\%    &  72.76\%  & 73.04\%   \\  
%orig: 66.23
& 73.44\% & 76.63\%    &  80.48\%  & \textbf{79.33}\%   \\ \hline
\multirow{1}{*}{\textbf{\begin{tabular}[c]{@{}c@{}}Incept. V3\end{tabular}}} 
%& 800  &  & 71.84\%    &  72.63\%  & 72.68\%   \\  
%& 1500 &  & 72.62\%    &  75.58\%  & 73.83\%   \\   
%orig: 63.00
& 72.10\%  & 74.20\%    &  76.42\%  & \textbf{74.33}\%   \\ \hline
\multirow{1}{*}{\textbf{\begin{tabular}[c]{@{}c@{}}DenseNet 121\end{tabular}}} 
%& 800  &  & 71.84\%    &  72.63\%  & 72.68\%   \\  
%& 1500 &  & 72.62\%    &  75.58\%  & 73.83\%   \\  
%orig:64.12
& 67.22\% & 71.29\%    &  75.22\%  & \textbf{72.33}\%   \\ \hline
\end{tabular}}
\label{table:retino}
\end{table}

\subsection{Diabetic Retinopathy Detection}
The task of Diabetic Retinopathy (DR) detection is to 
classify high-resolution retina imaging data collected from a Kaggle challenge\footnote{\url{https://www.kaggle.com/c/diabetic-retinopathy-detection}}.
%In this section, we focus on reprogramming network to perform classifying high-resolution retina data collected from kaggle. 
The goal is to predict different scales ranging from 0 to 4 corresponding to the rating of presence of DR.  Note that collecting labeled data for diagnosing DR is a costly and time-consuming process, as it requires experienced and well-trained clinicians to make annotations on the digital retina images. The collected dataset contains 5400 data samples and we hold 2400 data samples as the test set.
%So far, it takes long time for people to diagnose DR by assessing digital images of retina even a well trained clinician. Time-consuming process may lead to delayed treatment of patients.
In this task, we set the parameters $\eta=0.05$, $q=55$ and use 10 labels per target class for MLM. Table \ref{table:retino} shows the test accuracy of reprogramming different pretrained classifiers, including ResNet 50, Inception V3 and DenseNet 121. BAR can achieve 79.33\% accuracy, which is 2.7\% better than SOTA and nearly the same as white-box AR (80.48\%). We note that even without complicated techniques such as data augmentation and model emsemble, the performance of AR/BAR is close to the current best reported accuracy (81.36\%) in the literature~\cite{DR_Sota} using single model without ensemble approach, which requires specifically designed data augmentation with fine-tuning on ResNet 50.

%We use BAR to reprogram pretrained Resnet 50 and Inception-V3 ImageNet classifiers to perform DR detection with different training data sizes.\textcolor{magenta}{As shown in Table\ref{table:retino}, we evaluate with different training data size (3000/1500/800) and the testing data size is 2400. In this task, we set the mini batch size of each training epoch as 20, initial learning rate as 0.005 and parameter $q$ as 15. Same as previous experiments, we leverage MLM technique on DR task. For each 5 classes of DR, we map each of them with 15 Imagenet labels.} 

%We demonstrate that our result can achieve nearly performance as using general Convolutional neural network.

\subsection{Melanoma Detection}
Skin cancer is the most common type disease, with over 5 million newly diagnosed cases in the United States every year. However, visual inspection of the skin
and differentiating the type of skin diseases still remains as a challenging problem. ML-based approaches have been actively studied to address this challenge.
Here the target-domain dataset is extracted from the International Skin Imaging Collaboration (ISIC) \cite{codella2019skin, Tschandl2018} dataset, containing 10015 images of 7 types of skin cancer. The average image size is 450$\times$600 pixels. We resize these data samples to be 64$\times$64 pixels and embed them in the center of ImageNet-sized inputs. Since the data distribution is imbalanced (70\% data samples belong to one class), we perform re-sampling on the training data to ensure the same sample size for each class. Finally, the training/testing data samples are 7800/780. 
In this task, we assign 10 separate ImageNet labels to each target-domain label for MLM and set the parameters $\eta=0.05$ and $q=65$.
Table \ref{table:isic} reports the test accuracy of different methods.  Consistent with previous findings, BAR attains similar accuracy as AR. More importantly, their accuracy significantly increases the accuracy of finetuning by a significant margin (5-10\%), especially for Inception V3 and DenseNet 121 models. Training from scratch again suffers from insufficient data samples and hence has low accuracy. The performance of BAR/AR even outperforms the best reported accuracy (78.65\%) in the literature, which uses specifically designed data augmentation with finetuning on DenseNet \cite{li2018skin}.

%The accuracy of BAR is 81.71\% comparable to white-box AR accuracy 82.05\% and is also higher than the state-of-the-art accuracy 78.65\% on this task as reported in ISIC 2018 Challenge \cite{codella2019skin} which used single DenseNet model with data augmentation. We also note that transfer learning baseline is close to random guessing due to limited training data, where as BAR is 5-20\% better than the  transfer learning finetuned baseline.

\begin{table}[t]
\centering
%\footnotesize
%\setlength{\tabcolsep}{4pt}
\caption{Test accuracy on melanoma detection task. The notation $*$ denotes the network used in SOTA method.}
\vspace{1.5mm}
\adjustbox{max width=1\linewidth}{
\begin{tabular}{ccccc}
\hline
\textbf{Model}  & \textbf{\begin{tabular}[c]{@{}c@{}}From Stratch\end{tabular}} & \textbf{\begin{tabular}[c]{@{}c@{}}Finetuning\end{tabular}} & \textbf{\begin{tabular}[c]{@{}c@{}}AR\end{tabular}} & \textbf{\begin{tabular}[c]{@{}c@{}}BAR\end{tabular}} \\ \hline
\multirow{1}{*}{\textbf{\begin{tabular}[c]{@{}c@{}}ResNet 50\end{tabular}}} 
%orig: 59.10
& 72.10\% & 76.90\%    &  82.05\%  & \textbf{81.71}\%   \\ \hline
\multirow{1}{*}{\textbf{\begin{tabular}[c]{@{}c@{}}Incept. V3\end{tabular}}} 
%orig: 52.91
& 70.91\% & 68.63\%    &  82.01\%  & \textbf{80.20}\%   \\ \hline
\multirow{1}{*}{\textbf{\begin{tabular}[c]{@{}c@{}}DenseNet 121*\end{tabular}}} 
%orig: 52.28
& 70.22\% & 68.88\%    &  80.76\%  & \textbf{78.33}\%   \\ \hline
%\multirow{1}{*}{\textbf{\textbf{\begin{tabular}[c]{@{}c@{}}Densenet \\with Aug (sota).\end{tabular}}}} 
%& 72.74\% & 78.65\%    &    &    \\  
%& & & &   \\ \hline
\end{tabular}}
\label{table:isic}
\end{table}

\subsection{Reprogramming Real-life Prediction APIs}
To further demonstrate the practicality of BAR in reprogramming access-limited (black-box) ML models, we use two real-life online ML-as-a-Service (MLaaS) toolkits provided by Clarifai.com and Microsoft Custom Vision. For Clarifai.com, a regular user on an MLaaS platform can provide any data input (of the specified format) and observe a model's prediction via Prediction API but has no information about the model and training data used. For Microsoft Custom Vision, it allows users to upload labeled datasets and trains a ML model for prediction, but the trained model is unknown to users. We aim to show how BAR can ``unlock'' the inference power of these unknown ML models and reprogram them for Autism spectrum disorder classification or Diabetic retinopathy detection tasks. 
%Below we provide a brief introduction to two prediction APIs offered by Clarifai.com and discuss their reprogramming performance using BAR.  
Note that white-box AR and current transfer learning methods are inapplicable in this setting as acquiring input gradients or modifying the target model is inadmissible via prediction APIs.
%Nowadays, Machine Learning as a Service (MLaaS) Platform allows users to upload their well-labeled images to train models by using the built-in algorithms or directly use the trained models, whose training set is usually inaccessible to the users. For these services, users can access the API provided by MLaaS and obtain corresponding classification results with chosen inputs. 

\begin{table}[t]
%\footnotesize
%\setlength{\tabcolsep}{3pt}
\centering
\caption{Performance of BAR on Clarifai.com APIs.}
\vspace{1.5mm}
\adjustbox{max width=1\linewidth}{
\begin{tabular}{cccccccc}
\hline
\textbf{Orig. Task to  New Task} & \textbf{\begin{tabular}[c]{@{}c@{}}\textbf{ q}\end{tabular}} & \textbf{\# of query}  &\textbf{\begin{tabular}[c]{@{}c@{}}Accuracy\end{tabular}} & \textbf{Cost}     \\ \hline
\multirow{2}{*}{NSFW to  ASD} 
 & 15  & 12.8k  & 64.04\%  & \$14.24 \\
 & 25   & 24k   & \textbf{65.70}\%    & \$23.2  \\ \hline
\multirow{2}{*}{Moderation to  ASD} 
& 15   & 11.9k & 65.14\%    &  \$13.52 \\
& 25   & 23.8k  & \textbf{67.32}\%    &  \$23.04  \\ \hline
\multirow{2}{*}{Moderation to  DR} 
& 15   & 15.2k & 71.03\%    &  \$18.24 \\
& 25   & 26.4k & \textbf{72.75}\%    &  \$31.68
  \\ \hline
\end{tabular}}
\label{table:clarifai}
\vspace{-2mm}
\end{table}

\begin{table}[t]
\centering
\caption{Performance of BAR on Microsoft Custom Vision API.}
\vspace{1.5mm}
\adjustbox{max width=1\linewidth}{
\begin{tabular}{clllc}
\hline
\multicolumn{1}{c}{\textbf{Orig. Task to New Task}} & \textbf{q}  & \textbf{\# of query} & \textbf{Accuracy} & \multicolumn{1}{l}{\textbf{Cost}} \\ \hline
\multirow{3}{*}{\begin{tabular}[c]{@{}c@{}}Traffic sign classification \\ to\\  ASD\end{tabular}} 
& 1  & 1.86k  & 48.15\%  & \$3.72   \\
& 5  & 5.58k  & 62.34\%  & \$11.16   \\ 
& 10 & 10.23k & \textbf{67.80\%}  & \$20.46
\\ \hline
\end{tabular}}
\label{table:msapi}
\vspace{-2mm}
\end{table}

\textbf{Clarifai Moderation API.}
This API can recognize whether images or videos have contents such as ``gore'', ``drugs'', ``explicit nudity'', or ``suggestive nudity''. It also has a class called ``safe'', meaning it does not contain the aforementioned four moderation categories. Therefore, in total there are 5 output class labels for this API.

\textbf{Clarifai Not Safe For Work (NSFW) API.}
This API can recognize images or videos with inappropriate contents (e.g., ``porn'', ``sex'', or  ``nudity''). It provides the prediction of two output labels ``NSFW'' and ``SFW''. 
%\footnote{\url{https://www.clarifai.com/models/nsfw-image-recognition-model-e9576d86d2004ed1a38ba0cf39ecb4b1}}.
%Here, we use BAR to reprogram this API for ASD classification.

%Here we use BAR to reprogram these two API for ASD classification task. 
Here, we separate the ASD dataset into 930/104 and the DR dataset into 1500/2400 for training and testing, and in BAR we use random label mapping instead of frequency mapping to avoid extra query cost. % set $q=15$, use 1500/800 images for training and 2400 images for testing DR detection. Table \ref{table:clarifai} shows that BAR can achieve 72.75\% (71.03\%) test accuracy when using 1500 (800) training data samples.
The test accuracy, total number of queries and the expenses of reprogramming Clarifai.com are reported in Table \ref{table:clarifai}. For instance, to achieve 67.32\% accuracy for ASD task and 72.75\% for DR task, 
BAR only costs \$23.04 US dollars and \$31.68 for reprogramming the Clarifai Moderation API. Setting a larger $q$ value for a more accurate gradient estimation can indeed improve the accuracy but at the price of increased query and expense costs.
%Table \ref{table:clarifai} shows that BAR can achieve 67.32\% (65.14\%) accuracy when using 25 (15) $q$ random vectors for gradient estimation, which significantly outperforms the transfer learning baseline \$50.10. The estimated expense of renting an equivalent computing platform as our machine (Intel 9900K CPU with 1 NVIDIA RTX 2080 Ti GPU and 64 GB RAM) on Amazon Web Service (AWS) to train BAR only costs \$3.06 US dollars\footnote{The price of using AWS V100 p3.2xlarge instance for an hour}.
We expect the accuracy of BAR can be further enhanced if we use frequency-based multi-label mapping or reprogram prediction APIs with more source labels.
%In our implementation, we use one-to-one label mapping, set $q=25$, use 930/465 samples for training and 104 samples for testing. Table \ref{table:clarifai} shows that BAR can achieve 62.3\% (60.14\%) accuracy when using 930 (465) training data samples, which significantly outperforms the CNN baseline. With 930 training data samples, it only costs \$23.04 US dollars for Clarifai.com and \$3.06 US dollars for AWS. 
%\textcolor{blue}{and the total cost for reprogramming Clarifai NSFW API to perform ASD task with 62.30\% testing accuracy is around \$23.04 US dollars.}

\textbf{Microsoft Custom Vision API.}
 We use this API to obtain a black-box traffic sign image recognition model (with 43 classes) trained with GTSRB dataset~\cite{Stallkamp2012}. 
 %Note that the model is unknown to us as well. 
 We then apply BAR with different number of random vectors $q$ (1/5/10) and a fixed number of random label mapping $m=6$ to reprogram it for ASD task. As shown in Table \ref{table:msapi}, the test accuracy achieves 69.15\% when $q$ is set to 10 and the overall query cost is \$20.46 US dollars.

\begin{figure*}[t]
\centering
\includegraphics[width=0.99\linewidth]{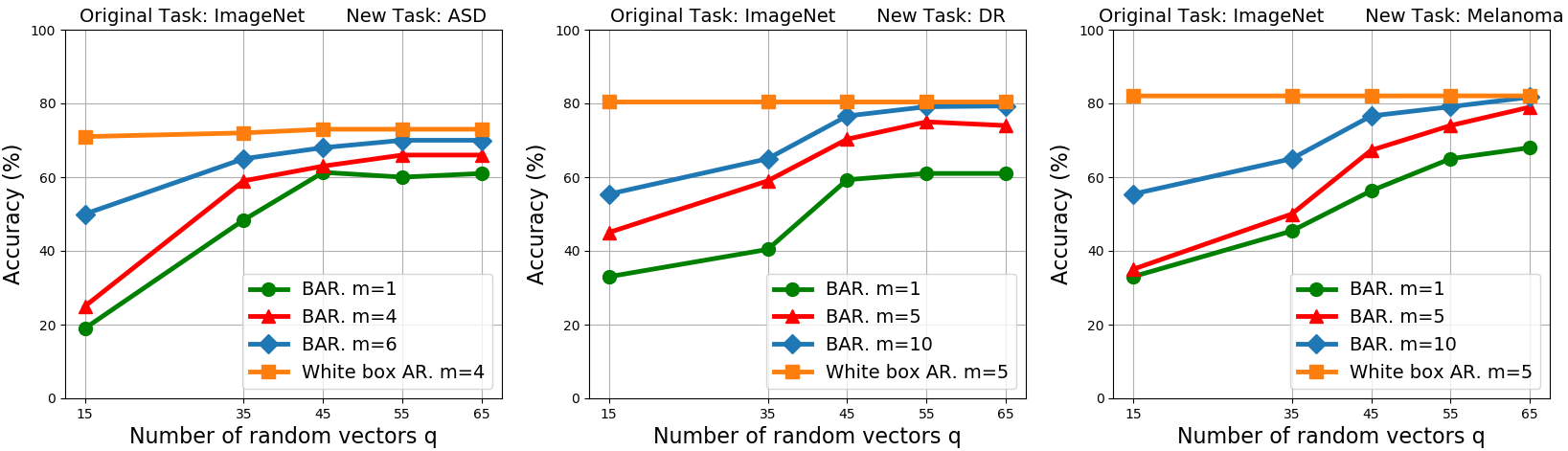}
%\vspace{-0.2in}
\vspace{-2mm}
\captionsetup{justification=centering}
\caption{Sensitivity analysis of BAR on the number of random vectors $q$ for gradient estimation and the frequency-based multi-label mapping size $m$.
The white-box AR is shown as a reference. The accuracy of BAR improves as $q$ or $m$ increases.}
\label{fig:mlm_q}
\vspace{-2mm}
\end{figure*}

\subsection{Ablation Study and Sensitivity Analysis}
\label{subsec_ablation}
%\begin{itemize}
\textbf{Number of random vectors ($q$) and mapping size ($m$).}
In our BAR method, we use the one-sided averaged gradient estimator in \eqref{eqn_avg_gradient_est} via $q$ random vector perturbations. 
%and we discuss that increasing $q$ will improve the gradient estimation but will also incur more model queries.
Here, we empirically investigate the sensitivity of $q$ and $m$ on the accuracy of BAR when reprogramming the pretrained Resnet 50 ImageNet model to perform ASD classification, DR detection,  and Melanoma detection with different $q$ and $m$ values.
%Noted that increasing $q$ can lead to more accurate direction in optimizaton. We empirically shows that number of $q$ directly affect the effectiveness of BAR. 
As shown in Figure \ref{fig:mlm_q}, the test accuracy of BAR is low when $q=15$, suggesting that insufficient gradient estimation will undermine the performance. 
On the other hand, the accuracy indeed increases with $q$ and then saturates for different mapping sizes. For a fixed $q$ number, we can conclude that increasing the label mapping size $m$ for each target-domain label can improve the accuracy.

%and mapping number $m$ for each 3 task due to better gradient estimation. For ASD task, the improvement of accuracy begins to saturate when $q=55$, achieving 72.99\% 10-fold test accuracy. Similarly, for DR and ISIC task, the accuracy curve becomes stable when $q=65$, achieving 79.13\% and 81.71\%.
%For ASD task, while $q$ increases to 25, the accuracy achieve 62.13\%.
%We also observe from Figure \ref{fig:mlm_q} that with a sufficient number of random directions for gradient estimation, BAR is able to maintain similar accuracy on these two social good tasks when the training data size is reduced, implying its effectiveness with limited data.

\begin{figure}[t]
\centering
\includegraphics[width=0.99\linewidth]{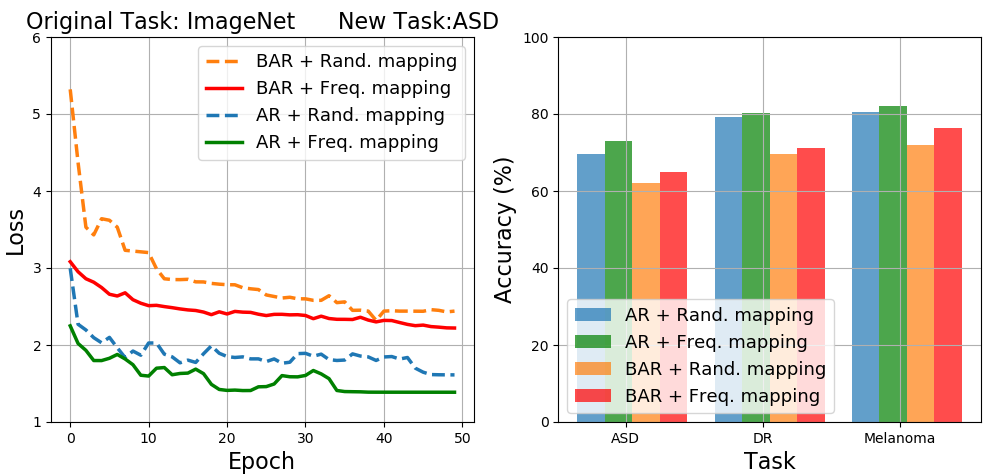}
%\vspace{-0.2in}
\vspace{-2mm}
\caption{Ablation study on random and frequency multi-label mapping for AR and BAR. With frequency mapping,
the accuracy and training performance of AR and BAR can be further improved.
Left: training loss over epochs for ASD task. Right: Test accuracy upon convergence for each task.
}
\label{fig:mapping}
%\vspace{-0.2in}
\end{figure}

\textbf{Random and frequency multi-label mapping.}
We perform an ablation study on two multiple-label mapping schemes -- random mapping and frequency mapping -- for both AR and BAR on Resnet 50 with the three medical imaging learning tasks. 
%The training data size for DR/ASD are 3000/930 samples.
For random mapping, for each target-domain class we randomly assign $m$ separate labels from the source domain. For frequency mapping, in each task, we first obtain the source-label prediction distribution of the target-domain data before reprogramming. Based on the distribution, we then sequentially assign the most frequent source-label to the corresponding dominating target-label until each target-label has been assigned with $m$ source-labels.
Figure \ref{fig:mapping} shows the training loss over training epochs (left diagram) on ASD and the resulting test accuracy (right diagram) upon convergence for all tasks. Comparing to random mapping, we find that frequency mapping leads to faster and better convergence results for both AR and BAR, thereby yielding roughly 3\% to 5\% gain in test accuracy. Similar trends in convergence are observed in DR and Melanoma detection tasks.

\begin{figure}[t]
\centering
\includegraphics[width=0.99\linewidth]{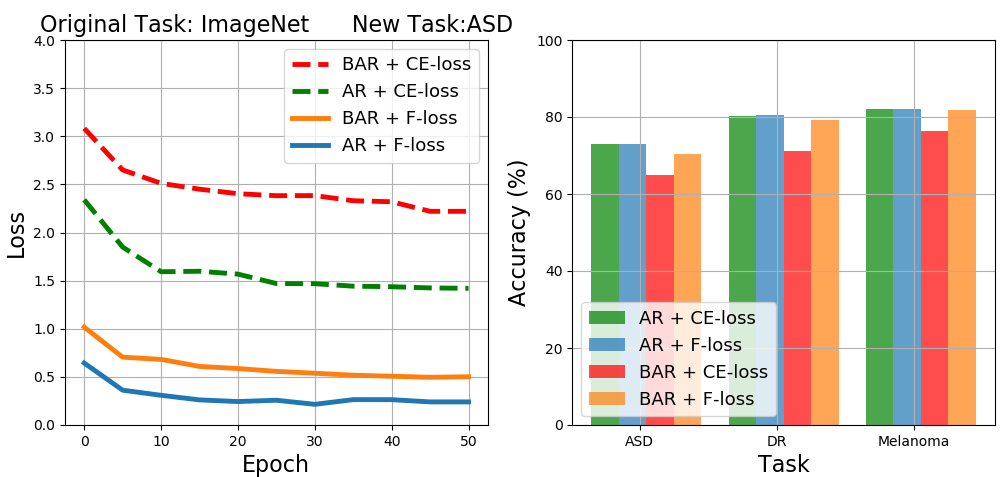}
%\vspace{-0.2in}
\vspace{-2mm}
\caption{Analysis on cross entropy (CE-loss) and focal loss (F-loss) for AR and BAR in each task. By using F-loss, the loss converges faster than using CE-loss for both AR and BAR methods. Using F-loss, the performance of BAR can be significantly improved in each task.}
\label{fig:focal}
%\vspace{-2mm}
\end{figure}

\begin{figure*}[t]
\centering
\includegraphics[width=0.99\linewidth]{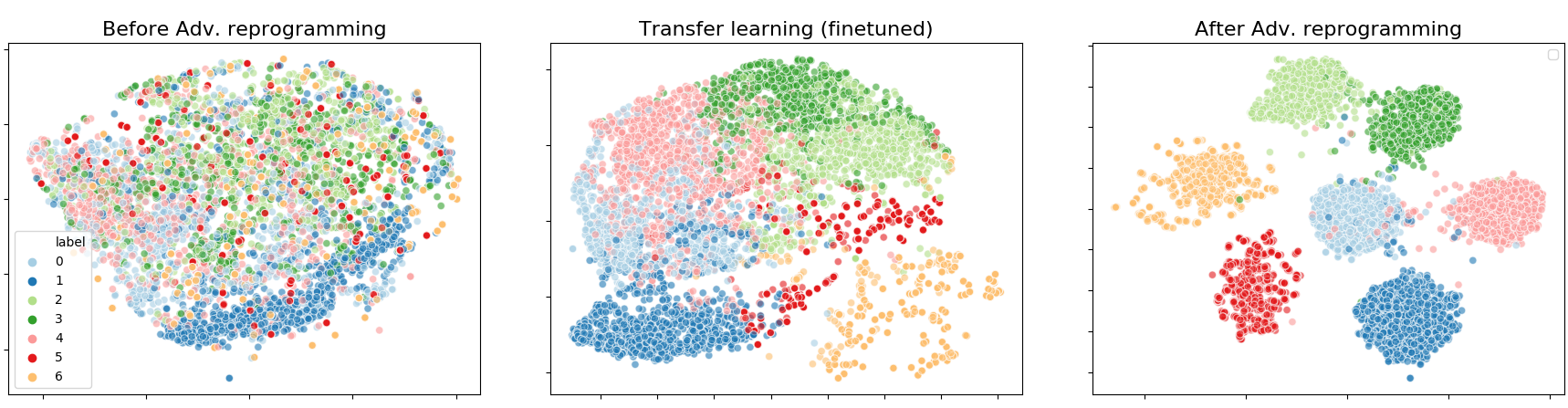}
%\vspace{-0.2in}
\vspace{-2mm}
\caption{Comparison of t-SNE embedding plots using ResNet 50 and the training data of Melanoma Detection task. Colors represent 7 different class labels. Adversarial reprogramming indeed learns better data representations for solving the target-domain task. }
\label{fig:tsne}
%\vspace{-0.2in}
%\vspace{-2mm}
\end{figure*}

\textbf{Cross entropy loss (CE-loss) and focal loss (F-loss).}
Here we compare the performance of AR and BAR using CE-loss and F-loss. As shown in Figure \ref{fig:focal} (left diagram), on ASD we find that using F-loss can converge faster and better than using CE-loss for both AR and BAR. Similar observations are made for DR and Melanoma tasks.
The performance gain when using F-loss can be explained by the fact that it is designed for improving dense object detection, with the capability of better differentiating foreground-background variances, which well maps to our AR setting (foreground being the embedded target-domain data and background being the learned universal adversarial program). Comparing the test accuracy on different tasks (right diagram), we find that F-loss greatly improves the accuracy of BAR by 3\%-5\%, while the gain in AR's accuracy can be marginal.

\textbf{Representation analysis.} To validate that AR/BAR indeed learns useful data representations for transfer learning, in Figure \ref{fig:tsne} we visualize the data representations of before/after AR and finetuning using t-distributed stochastic neighbor embedding (t-SNE), where the melanoma training data representations  are extracted from the ResNet 50 feature maps of the pre-logit layer. We can observe that before AR, the data representations are non-separable, whereas after AR they become highly clustered and well separated, leading to high predictability. In contrast, finetuning has worse representation learning performance relative to AR. The t-SNE plots of other datasets are shown in the supplementary material.

%In ASD task, we set $\alpha=0.7$, $\gamma=2$ for AR and BAR (Parameter for other tasks show in appendix). The accuracy of BAR with focal loss in ASD task can achieve 70.33\%, which is better than CE-loss 65.00\%. Overall, using focal loss improves 
%$\sim5\%$ to $\sim8\%$ performance in each task. 
%\end{itemize}
%\begin{figure}[t]
%\centering
%\includegraphics[width=0.97\linewidth]{figs/tsne_adv_tr%ain2.png}
%\vspace{-0.2in}
%\caption{tsne after reprogramming}
%Ablation study on multi-label mapping (MLM). With MLM, %the accuracy of BAR can be further enhanced.}
%\vspace{-0.2in}
%%\label{fig:multli_label_mapping}
%\end{figure}

%\begin{table}[]
%\footnotesize
%\setlength{\tabcolsep}{4pt}
%\begin{tabular}{cccccccc}
%\hline
%\textbf{Task} & \textbf{training size} & \textbf{testing size}  & %\textbf{\# of query} & \textbf{accuracy} & \textbf{cost}     \\ \hline
%\multirow{2}{*}{DR} 
%& 800  & 2400  & 12.8k   & 72.67\%  & \$14.24 \\
%& 1500 & 2400  & 24k     & 73.57\%  & \$23.2  \\ \hline
%\multirow{2}{*}{ASD} 
%& 459  & 117   & 11.9k   & 60.59\%    &  \$13.52 \\
%& 918  & 117   & 23.8k   & 62.10\%    &  \$23.04  \\ \hline
%\end{tabular}
%\end{table}

%% file: doc/conclusion.tex
In this paper, we proposed BAR, a novel approach to adversarial reprogramming of black-box ML models via zeroth order optimization and multi-label mapping techniques. Comparing to the vanilla AR method assuming complete knowledge of the target ML model, our BAR method only required input-output model responses, enabling black-box transfer learning of access-limited ML models. Evaluated on three data-scarce medical ML tasks, BAR showed comparable performance to the vanilla white-box AR method and outperformed the respective state-of-the-art methods as well as the widely used finetuning approach. We also demonstrated the practicality and effectiveness of BAR in reprogramming real-life online image classification APIs with affordable expenses, and performed in-depth ablation studies and sensitivity analysis. Our results provide a new perspective and an effective approach for transfer learning without knowing or modifying the pre-trained model.
%Our future work includes extending our BAR framework to ML tasks beyond classification.

% \textcolor{blue}{
% \begin{enumerate}
%     \item Deal with input
% \end{enumerate}
% }

%% file: doc/supplementary.tex
\section*{Supplementary Material}

\section{Training details}
\subsection{Transfer learning (finetuned): }
For fine-tuning, our implementation  follows the TensorFlow tutorial\footnote{\url{https://www.tensorflow.org/tutorials/images/transfer_learning}}. By using the Keras package in TensorFlow,
%We use the KerasNet as our basic architecture, which is a high level network with pretrained weights in Tensorflow, 
we can transfer-learning a pretrained Keras network for new classification tasks. Since in our case, the target domains of the datasets are quite different from the source domain of the pretrained model (ImageNet dataset). We use the convolutional layers as our base model. Additionally, we add a fully connected layer following with a dropout layer, and we set the dropout rate to 0.5 before the last classification softmax layer. We find that finetuning every layer performs better than just finetuning the last fully connected layer.
\begin{itemize}

\item In Autism Spectrum Disorder (ASD) task, we fine-tune ResNet 50 and Inception V3 using the same pretrained networks as adversarial reprogramming in Section 4. Here, we use Adam with learning rate of 1e-5 as the optimizer and set the batch size to 32. The training epoch is set to 50.

\item In Diabetic Retinopathy (DR) and Melanoma detection tasks, we follow the training setting in ASD task and fine-tune three networks, including ResNet 50, Inception V3 and DenseNet 121.
%we add additional layers and do not include the top final layer of the original networks.
\end{itemize}
\subsection{Training from scratch: }
Following the same architectures as finetuning, we train the networks with random initializer on ASD, DR and Melanoma datasets instead of using any pretrained weights. We use Adam as optimizer and set the learning rate to 1e-5 with the batch size of 32. The training epoch is set to 50.

\section{Additional experiments}
We conducted additional experiments on reprogramming pre-trained ImageNet models for the facial recognition task on Georgia Tech Face Database\footnote{\url{https://computervisiononline.com/dataset/1105138700}} (GTFD, 50 classes with 15 face images/class for training and 150 images for testing). The results are shown in Table \ref{table:face}. Similar to other tasks show in Section \ref{sec:experiment}, BAR performs much better than baseline methods and is comparable to AR.

\begin{table}[t]
\centering
\caption{Test accuracy on facial recognition task.}
\adjustbox{max width=1\linewidth}{
\begin{tabular}{ccccc}
\hline
\textbf{Model}        & \textbf{From scratch} & \textbf{Finetuning} & \textbf{AR} & \textbf{BAR}     \\ \hline
\textbf{ResNet 50}    & 95.90\%               & 97.67\%             & 99.29\%     & \textbf{98.32\%} \\ \hline
\textbf{Inpcept. V3}  & 92.11\%               & 95.33\%             & 97.66\%     & \textbf{97.36\%} \\ \hline
\textbf{DenseNet 121} & 90.12\%               & 91.20\%             & 97.25\%     & \textbf{95.90\%} \\ \hline
\end{tabular}}
\label{table:face}
\end{table}

\section{t-SNE}
 We further visualize data representation of the ASD and DR datasets using t-Distributed Stochastic Neighbor Embedding (t-SNE), a tool to visualize high-dimensional data. We use open-source Scikit-Learn library to implement t-SNE with general perplexity of 50.
 
For every data point, we extract the hidden representations from the pre-logit layer of ResNet 50 on three datasets, illustrating the difference among "before adversarial reprogramming (AR)", "after AR" and "transfer learning (fine-tuned)". First, for before AR, we pad training data with zero values to fit the input size of source domain. Second, we extract the feature values of input data from transfer learning model with finetuning. Third, for the feature after adversarial reprogramming, we extract the values of original data with the adversarial program.

\subsection{Autism Spectrum Disorder classification}
\begin{figure*}[t]
\centering
\includegraphics[width=1\linewidth]{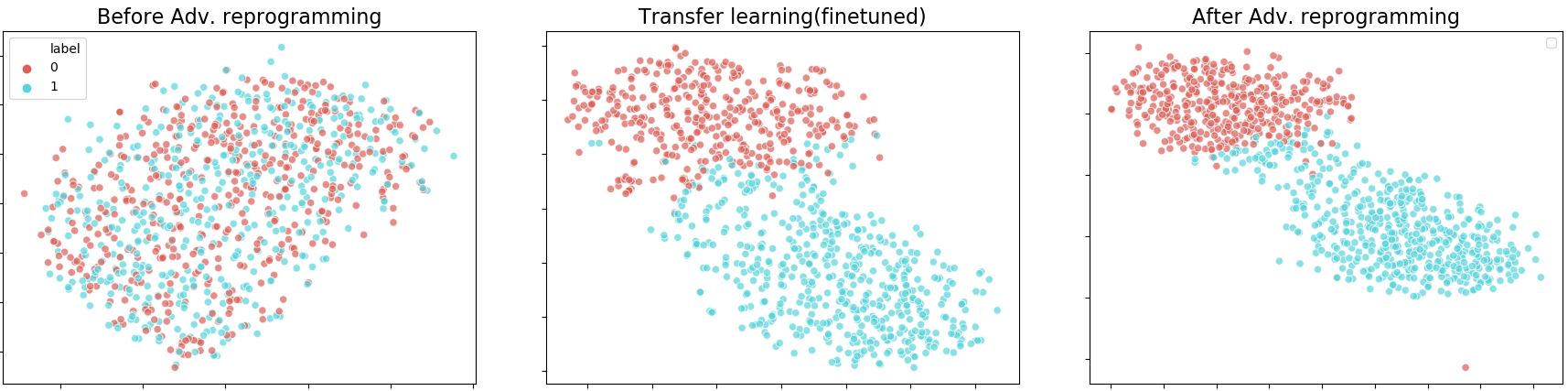}
%\vspace{-0.2in}
\caption{Comparison of t-SNE representations using ResNet 50 and the training data of ASD classification task.  Colors represent 2 different class labels. }
\label{fig:asd_tsne_orig}
%\vspace{-0.2in}
\end{figure*}

As shown in Figure \ref{fig:asd_tsne_orig}, we visualize the data representations of before/after AR and finetuning using t-distributed stochastic neighbor embedding (t-SNE). Colors represent 2 different class labels (ASD/Non-ASD). We can observe that before AR, the data representations are non-separable, whereas after AR they become highly clustered and well separated, leading to high predictability. In contrast, finetuning has worse representation learning performance relative to AR. Adversarial reprogramming indeed learns better data representations for solving the target-domain task.  

\subsection{Diabetic Retinopathy detection}
\begin{figure*}[t]
\centering
\includegraphics[width=1\linewidth]{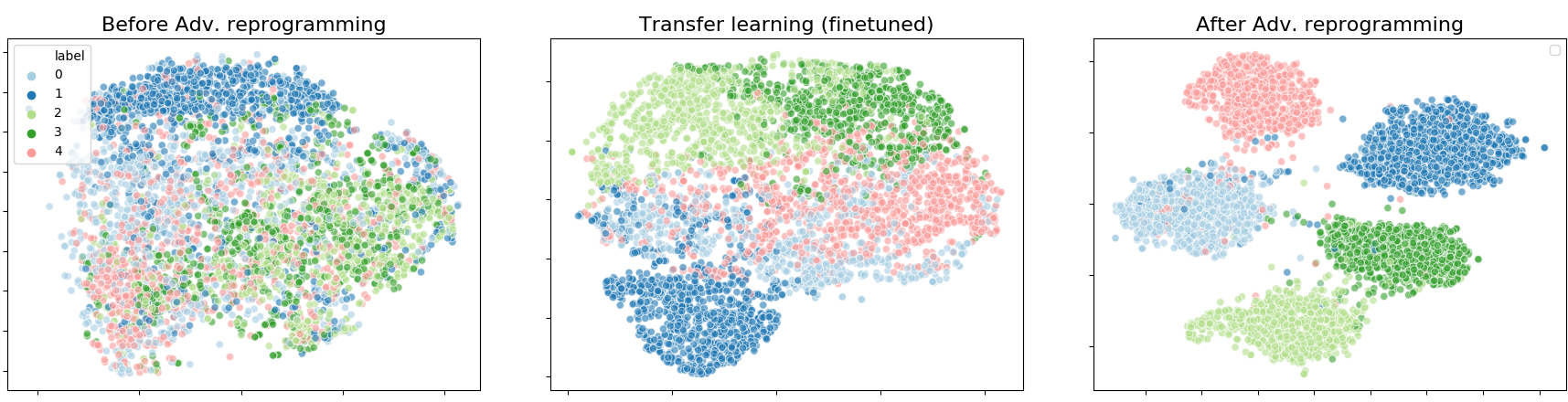}
%\vspace{-0.2in}
\caption{Comparison of t-SNE representations using ResNet 50 and the training data of DR Detection task. Colors represent 5 different class labels.}
\label{fig:dr_tsne_orig}
%\vspace{-0.2in}
\end{figure*}

 The t-SNE plot is shown in  Figure \ref{fig:dr_tsne_orig}.  Adversarial programming learns better data representation than finetuning.

\subsection{Melanoma detection}
\begin{figure*}[h]
\centering
\includegraphics[width=1\linewidth]{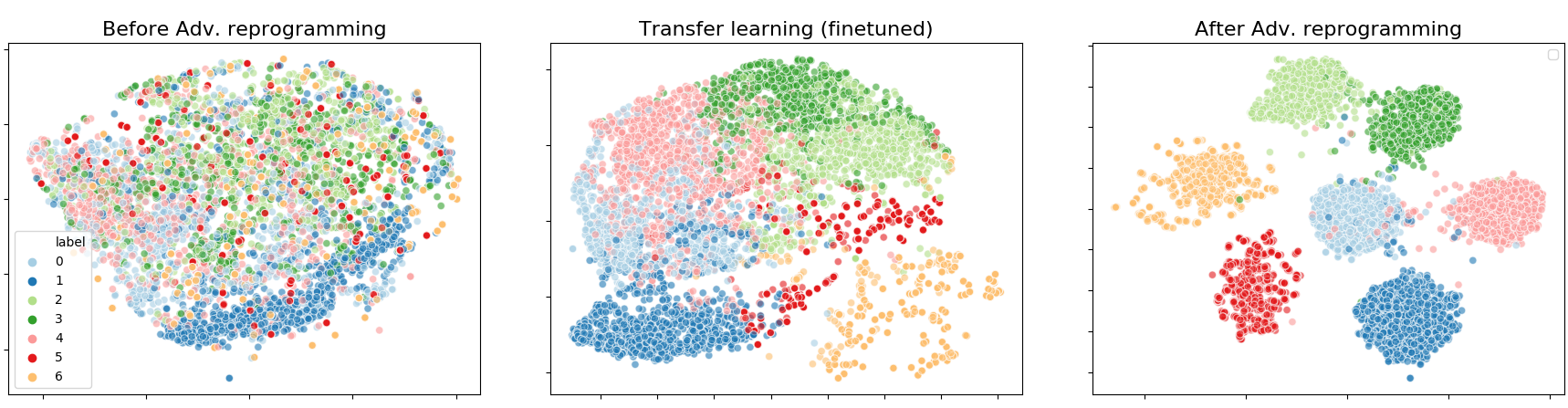}
%\vspace{-0.2in}
\vspace{-2mm}
\caption{Comparison of t-SNE representations using ResNet 50 and the training data of Melanoma Detection task. Colors represent 7 different class labels. }
\label{fig:tsne_isic}
%\vspace{-0.2in}
\vspace{-2mm}
\end{figure*}

 The t-SNE plot is shown in  Figure \ref{fig:tsne_isic}. Consistent with the t-SNE plots of ASD and DR tasks, adversarial programming learns better data representation.

